\documentclass[conference]{IEEEtran}

\usepackage{times}
\usepackage{epsfig}
\usepackage{graphicx}
\usepackage{amsmath}
\usepackage{amssymb}
\usepackage{tablefootnote}
\usepackage[referable]{threeparttablex}
\usepackage{bm}
\usepackage{graphics} 
\usepackage{multirow}
\usepackage{pifont}
\usepackage{calc}
\usepackage{amsfonts} 
\usepackage[caption=false]{subfig}

\usepackage{amsmath} 
\usepackage[super]{nth}
\usepackage{balance}
\usepackage[pagebackref=true,breaklinks=true,letterpaper=true,colorlinks,bookmarks=false]{hyperref}
\newcommand{\onedot}{.\ }
\newcommand{\eg}{\emph{e.g.}}
\newcommand{\etal}{\emph{et al}\onedot}

\ifCLASSINFOpdf
\else
\fi

\newcommand{\Fig}[1] {Figure \ref{fig:#1}}

\newcommand{\Tbl}[1]  {Table \ref{tbl:#1}}

\def\eg{\emph{e.g}\onedot} 
\def\ie{\emph{i.e}\onedot}

\def\etal{\emph{et al}\onedot}

\newcommand{\cmark}{\ding{51}}%
\newcommand{\xmark}{\ding{55}}%

\begin{document}
%
\title{360$^\circ$ in the Wild: Dataset for Depth Prediction and View Synthesis}

\author{\IEEEauthorblockN{Kibaek Park}
\IEEEauthorblockA{
KAIST\\
School of EE\\
parkkibaek@kaist.ac.kr\\}
\and
\IEEEauthorblockN{Francois Rameau}
\IEEEauthorblockA{KAIST\\School of EE\\frameau@kaist.ac.kr}
\and
\IEEEauthorblockN{Jaesik Park}
\IEEEauthorblockA{SNU\\School of CSE\\
jaesik.park@snu.ac.kr}
\and
\IEEEauthorblockN{In So Kweon}
\IEEEauthorblockA{KAIST\\School of EE\\iskweon77@kaist.ac.kr}}


%


\maketitle
\IEEEpeerreviewmaketitle
\begin{abstract}
The large abundance of perspective camera datasets facilitated the emergence of novel learning-based strategies for various tasks, such as camera localization, single image depth estimation, or view synthesis. However, panoramic or omnidirectional image datasets, including essential information, such as pose and depth, are mostly made with synthetic scenes. In this work, we introduce a large scale 360$^{\circ}$ videos dataset in the wild. This dataset has been carefully scraped from the Internet and has been captured from various locations worldwide. Hence, this dataset exhibits very diversified environments (e.g., indoor and outdoor) and contexts (e.g., with and without moving objects). Each of the 25K images constituting our dataset is provided with its respective camera's pose and depth map. We illustrate the relevance of our dataset for two main tasks, namely, single image depth estimation and view synthesis. 
\end{abstract}


 \begin{figure}[t]
     \centering
     \includegraphics[width=.99\linewidth]{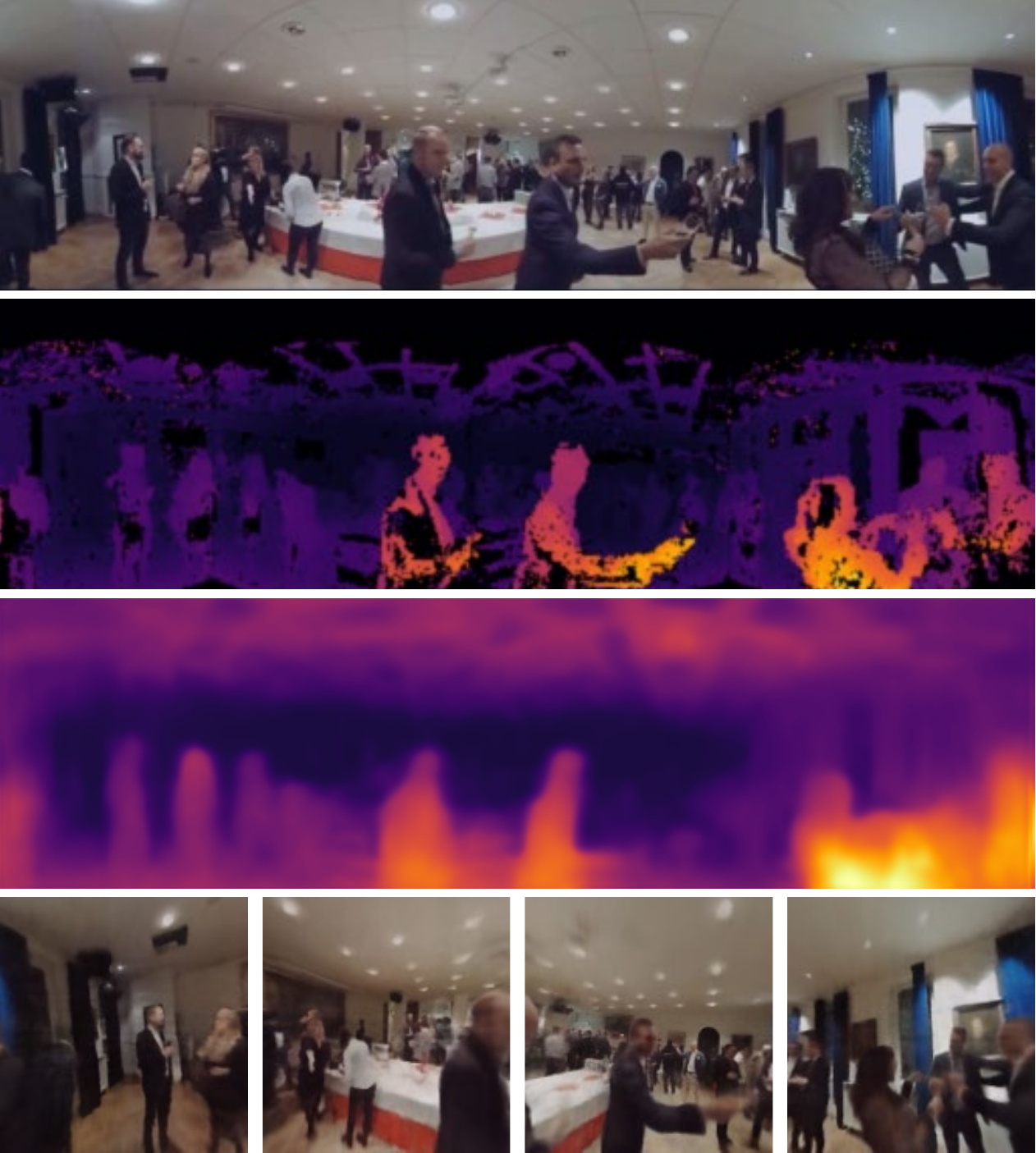}
     \caption{An example data of 360$^\circ$ in the Wild. The proposed dataset consists of omnidirectional videos, camera trajectories, successive scene depth. The dataset enables depth estimation and novel view synthesis with real-world videos. The figure shows omnidirectional RGB, scene depth, depth prediction by~\cite{Hu2019RevisitingSI}, and novel view synthesis represented as perspective views by using extended NeRF++~\cite{nerfpp}, from the top row to bottom, respectively. Images are cropped for the best view.
      }
     \label{fig:teaser}
 \end{figure}

%

\section{Introduction}

Large scale image datasets have played a key role in the development of modern deep learning networks. One striking example is the considerable impact of ImageNet~\cite{deng2009imagenet} on the machine learning community. Subsequently, many more datasets have been publicly released for various tasks. For instance, the KITTI~\cite{geiger2012we} dataset provides images, camera pose and 3D data for a wide spectrum of applications, such as, depth completion~\cite{park2020non}, visual odometry~\cite{mur2017orb} or 2D/3D object detection~\cite{choe2019segment2regress}. While all these datasets have undoubtedly contributed to the community, most of them have only been acquired by perspective cameras. Unfortunately, regardless of the application, the networks trained with such datasets are often incompatible with images exhibiting strong geometric distortions -- such as omnidirectional (or 360$^\circ$) images.   

With the rapid democratization of consumer-level omnidirectional cameras (e.g., Insta360, Ricoh Theta, and GoPro Max, to cite just a few), the urge for omnidirectional large scale dataset is pressing to develop ad hoc deep learning solutions.
Moreover, the need for omnidirectional vision systems is on the rise as it finds applications in various fields, such as immersive augmented reality~\cite{teo2019mixed}, vehicular navigation~\cite{zhang2016benefit}, robotics~\cite{rituerto2010visual} or video surveillance~\cite{rameau2011tracking}.
For these particular tasks, accurate camera pose and ground truth depth maps are desirable. Indeed, this information is useful to train supervised approaches and serve as a benchmark. Despite a clear need for a realistic $360^\circ$ image dataset, most existing datasets, including ground-truth pose and depth information, have been synthetically generated~\cite{karakottas2019360surface, zhang2016benefit, zioulis2019spherical}. The resulting synthetic images often lack realism and diversity, leading to well-known generalization problems~\cite{pan2020unsupervised}.

On the contrary, the datasets \emph{in the wild}~\cite{frozen_data, lou2019veri, soomro2012ucf101} are usually more diversified and realistic as they contain arbitrary environments, exhibit complex cameras trajectories, and have been acquired from various types of cameras. These datasets are often collected from video-sharing or photo hosting online platforms. A representative example is the Mannequin Challenge Dataset~\cite{frozen_data}, which was released to resolve the poor generalization of deep learning networks on moving objects (in particular on persons). Specifically, \cite{frozen_data} targets applications like single image depth estimation and view synthesis in dynamic environments. 

Inspired by this effort, we propose a novel \emph{in the wild} dataset specifically designed for \emph{omnidirectional images}\footnote{Download link will be released.}. 
Our dataset is composed of more than 25K images with corresponding depth maps and camera poses obtained by Structure from Motion~(SfM)~\cite{opensfm} and Multi-View Stereo~(MVS)~\cite{schoenberger2016mvs} methods. The images have been extracted from YouTube videos and contain diverse scenes and contexts.
To highlight the usefulness and practicability of the new dataset, we showcase two examples of applications: single image depth estimation~\cite{bifuse} and novel view synthesis~\cite{nerf, nerfpp}. Regarding the case of novel view synthesis, we adapted NeRF++\cite{nerfpp} to spherical cameras.
\\



Our contributions can be summarized as follows:
\begin{itemize}
    \item We introduce a new omnidirectional image dataset extracted from a large number of online videos. The dataset covers various classes of real-world scenes. 
    \item For each frame we provide their pose and depthmap computed from a SfM pipeline. Additionally, we make available binary masks to remove moving objects from the video sequences.
    \item We demonstrate our dataset's applicability through two applications: single image depth estimation and novel view synthesis.
\end{itemize}

\section{Related Works}

In the past decades, omnidirectional vision has been a fertile ground for research. In particular, 3D vision tasks from $360^{\circ}$ cameras such as Simultaneous Localization and Mapping (SLAM)~\cite{omni_lsdslam,czarnowski2020deepfactors,li2019pose,tateno2017cnn,zhou2020deeptam}, monocular depth estimation~\cite{huang20176,lee2019learning,lee2019instance,ranjan2019collaboration,zhou2017unsupervised} and multi-view stereo~\cite{chen2019point,huang2018deepmvs,Im-2018-110411,lai2019real,won2019omnimvs,yao2018mvsnet} have been extensively investigated.
However, in contrast with perspective data, only few datasets dedicated to omnidirectional vision have been publicly released~\cite{immersive,matterport3d,dai2017scannet,Ranftl2020,schops2017multi,gibson}.

This section reviews relevant tasks (depth estimation and view synthesis) and discuss existing omnidirectional datasets with their limitations.

\subsection{Tasks}

\vspace{1mm} \noindent \textbf{360$^{\circ}$ Depth Estimation} is a task of primary importance for various applications such as, SLAM, object detection and AR/VR.
While single image depth estimation approaches have been intensely investigated for perspective cameras, it attracted comparatively less attention for the case of omnidirectional cameras. The lack of readily available datasets for this task is undoubtedly one reason for this narrow literature. 

Depth can be obtained from a single 360$^\circ$ image by fully-supervised and self-supervised methods. The strategies belonging to this first category~\cite{bifuse,zeng2020joint,omnidepth} are trained using direct supervision from ground-truth data. In contrast, self-supervised methods use photometric consistency across successive views from a monocular video as a training signal~\cite{chen2020distortion,payen2018eliminating,wang2018self,zioulis2019spherical}. While self-supervised approaches are significantly less data-dependent, they suffer geometrical inconsistency if moving objects appear in the training image set. As demonstrated in the mannequin challenge dataset~\cite{frozen_data}, training data containing moving objects is required to ensure a suitable generalization of the network to general dynamic scenes.

In the context of single image depth estimation, our dataset is particularly pertinent. Specifically, it can be used for single image depth training and self-supervised depth estimation (thanks to the sequential nature of our data). 


\vspace{1mm} \noindent \textbf{Novel View Synthesis.} From one or multiple images, novel-view synthesis consists in generating new images that render the scene from an arbitrary viewpoint. 
Recently, NeRF~\cite{nerf} introduces an approach using the neural radiance field in each camera ray vector. Specifically, the colors of novel views are derived from the integration of the sampled radiance field. 
Despite compelling results, NeRF can hardly deal with large scenes due to the property of the normalized device coordinates (NDC). To tackle this limitation, NeRF++~\cite{nerfpp} proposes a new geometrical representation called inverted sphere parameterization to deal with this problem. 
On the other hand, NeRF-W~\cite{nerfw} improved the result by learning the uncertainty and transient object from internet photos in an unsupervised way. NeRF-W handles the dark fog problem of the NeRF occurred by the occluded objects and variable illumination changes. Note that all of the studies about the neural radiance field are demonstrated with perspective camera models only. 

While perspective image datasets providing accurate camera pose are abundant~\cite{Geiger2012CVPR,dai2017scannet,Knapitsch2017}, it is not the case for omnidirectional images. Thus, this dataset constitutes a solid benchmark for the development and testing of view synthesis techniques. Moreover, the variety of our data contains relevant challenges for view synthesis, such as large environments and complex indoor scenes. Finally, the large number of ``frozen people" in the dataset makes it applicable to scene containing moving persons.
In our study, we trained on omnidirectional images and showed the network's expandability through the proposed dataset.

\subsection{Datasets}

\noindent \textbf{Internet Videos} provide a variety of examples that can help learning-based approaches to be taught from various scenes.
For example, MegaDepth~\cite{megadepth} released the internet photo dataset for single view depth estimation, improving the generalization not only to MegaDepth data but also to other datasets ~(\eg~KITTI). Similarly, for relative depth estimation, DIW~\cite{diw}, ReDWeb~\cite{redweb} and YouTube3D~\cite{youtube3d} datasets help to increase the accuracy for arbitrary input image. Meanwhile, Li~\etal~\cite{frozen_data} concentrated on moving objects leading to failure on the image-based 3D reconstruction. The dataset consists of internet videos where the camera moves between stationary persons and the structure. The network trained with this dataset precisely estimates scene depth, including moving objects.

Obviously, the Internet videos taken by anonymous users contribute to the network generalization compared with most public datasets taken in restricted places and situations. 

\vspace{1mm} \noindent \textbf{360$^{\circ}$ datasets.} 
Several datasets provide 360$^{\circ}$ images for various purposes. For instance, accurate and dense 3D reconstructions of real indoor environments have been made available in the Stanford2D3D~\cite{2D-3D-Semantics} and Matterport3D~\cite{matterport3d} datasets. These annotated data are commonly used to train depth prediction networks with real images \cite{garcia2017review}.
Concurrently, synthetic datasets have been released~\cite{InteriorNet,wang2018self} for SLAM, optical flow, and dense depth estimation. Synthetic data has the advantage of providing a dense and accurate ground-truth, but their lack of realism often causes generalization issues~\cite{maximov2020focus}. 
In order to increase the transferability of approaches trained on 360$^{\circ}$ images, the dataset VCL-360~\cite{omnidepth,zioulis2019spherical,karakottas2019360surface} proposes to merge a collection of existing datasets (containing both real and synthetic images).
It should, however, be noted that this dataset exclusively covers indoor environments.
Recently, Maugey~\etal presented FTV360~\cite{maugey2019ftv360}, which contains indoor and outdoor sequences captured with 40 panoramic cameras. However, this dataset is mainly designed for novel view synthesis and is not suitable to train depth estimation architectures due to the limited number of scenes it contains.
Similarly, a hemispherical rig with a 46 wide-FOV action camera array was produced by Broxton~\etal~\cite{immersive}. It leverages the MPI representation to render view-dependent immersive light field video. Unfortunately, this dataset has not been released publicly.
Besides, Caruso~\etal~\cite{omni_lsdslam} carried out omnidirectional SLAM and released a small test-bed dataset composed of greyscale fisheye images.
Finally, omnidirectional images have also been generated from densely 3D scanned indoor environments~\cite{habitat,gibson,replica}. However, they usually contain indoor scenarios only and exhibit a limited variability of scenes and scenarios (see~~\Tbl{table_related_work}).

While the previously mentioned datasets allow to design and test novel network architectures, their lack of diversity restricts their usage to specific environments. To ensure the deployment of single image depth estimation to any arbitrary scene, we propose to introduce the \emph{first} internet-based 360$^{\circ}$ video dataset with pose and depth annotations. This new dataset is a collection of 273 videos captured from various scenes with arbitrary camera movements. 
To illustrate the usefulness and relevance of this new dataset, we trained and tested state-of-the-art single-view depth estimation and novel view synthesis approaches.

\begin{table*}[]
\renewcommand{\arraystretch}{1}
\centering
\caption{Comparison with existing of 360$^\circ$ datasets and proposed dataset.}
\begin{threeparttable}
\resizebox{0.92\linewidth}{!}{
\begin{tabular}{c||c|c|c|c|c|c|c|c|c}
\hline
\multirow{2}{*}{ Datasets} &
  \multirow{2}{*}{ Total images} &
  \multicolumn{4}{c|}{ 360$^{\circ}$ Scene} &
  \multicolumn{2}{c|}{ 360$^{\circ}$ Depth} &
  \multirow{2}{*}{\shortstack{ Camera \\  trajectory}} &
  \multirow{2}{*}{\shortstack{ Internet \\  videos}} \\ 
  \cline{3-6} \cline{7-8} 
                 &     &  Indoor &  Outdoor &  Real &  Synthetic &  num &  type &  &          \\ 
\hline
\hline
PanoSUNCG~\cite{wang2018self}             & 25K  & \cmark  & &  & 25K    &  25K    & \textit{Syn} & \cmark & \xmark  \\ \hline
InteriorNet~\cite{InteriorNet}     & 20M  & \cmark  &   &   & 20M$^1$ & --    & \textit{Syn} & \cmark & \xmark  \\ \hline
Stanford2D3D~\cite{2D-3D-Semantics} & 70K & \cmark &  &  70K &  & 70K    & \textit{ToF} & \xmark & \xmark  \\ \hline
Matterport3D~\cite{matterport3d}     & 10K  & \cmark &  & 10K  &  & 10K    & \textit{ToF}  & \xmark    & \xmark   \\ \hline
Payen~\etal~\cite{payen2018eliminating}             & 8K  &   & \cmark &   & 8K  & 200 & Syn & \xmark  & \xmark  \\ \hline
FTV360~\cite{maugey2019ftv360}             & 255K  & \cmark & \cmark & 255K$^2$  &   & -- & -- & \xmark  & \xmark  \\ \hline
\hline
Ours                                      & 25K & \cmark  & \cmark  &  25K  &  & 11K$^3$ & \textit{Mono} & \cmark & \cmark \\ \hline
\end{tabular}
}
\begin{tablenotes}
\begin{small}
    \item[*]Note that depth ground-truth data from synthetic data~(\textit{Syn}), time of flight devices~(\textit{ToF}) and SfM$+$MVS~(\textit{Mono}).
    \item[1] InteriorNet~\cite{InteriorNet} only provides perspective depth ground-truth.
    \item[2] FTV360~\cite{maugey2019ftv360} contains only three sequences acquired with 40 static cameras.
    \item[3] Occasionally, the 3D reconstruction fails to provide a reliable depth map. We reject these samples from consideration for depth estimation.
\end{small}
\end{tablenotes}
\end{threeparttable}
\label{tbl:table_related_work}
\end{table*}

\section{360$^\circ$ in the Wild Dataset}
In this section, we precisely describe the procedure we deployed to build our dataset. 
\subsection{Ground-truth Annotation}
Our dataset provides the camera's pose as well as the corresponding depth map for each image.
While many 3D reconstruction softwares~\cite{schonberger2016structure,opensfm,moulon2016openmvg} are currently available, only a few are compatible with omnidirectional vision systems. After an extensive comparison between candidate solutions, we opted for OpenSfM~\cite{opensfm} to perform the 3D reconstruction task in our sequences. Despite its robustness, OpenSfM remains very sensitive to the presence of moving objects in the scene. Thus, we searched 360$^\circ$ YouTube videos with the following keywords: \textit{pandemic lockdown, virtual tour, virtual open house, and mannequin challenge} and carefully selected videos. 

Our dataset provides the camera's pose as well as the corresponding depth map for each image. While many 3D reconstruction softwares~\cite{schonberger2016structure,opensfm,moulon2016openmvg} are currently available, only a few are compatible with omnidirectional vision systems. After an extensive comparison between candidate solutions, we opted for OpenSfM~\cite{opensfm} to perform the sparse mapping and camera pose estimation tasks in our sequences. OpenSfM~\cite{opensfm} also proposes a dense multi-view stereo reconstruction process, however, this implementation led to very sparse and noisy depth maps that are not suitable to train depth estimation networks. 
Therefore, to ensure a high quality depth pseudo labels, we utilize COLMAP~\cite{schoenberger2016mvs} which is known for its accuracy~\cite{frozen_data}. However, COLMAP is not compatible with 360$^{\circ}$ images, therefore, we transform the panoramic images into cube maps which are composed of 6 perspective images. This set of perspective images and their respective poses are then used in~\cite{schoenberger2016mvs} to compute dense and reliable depth maps.



\begin{figure}[tb]
    \centering
    \includegraphics[width=.99\linewidth]{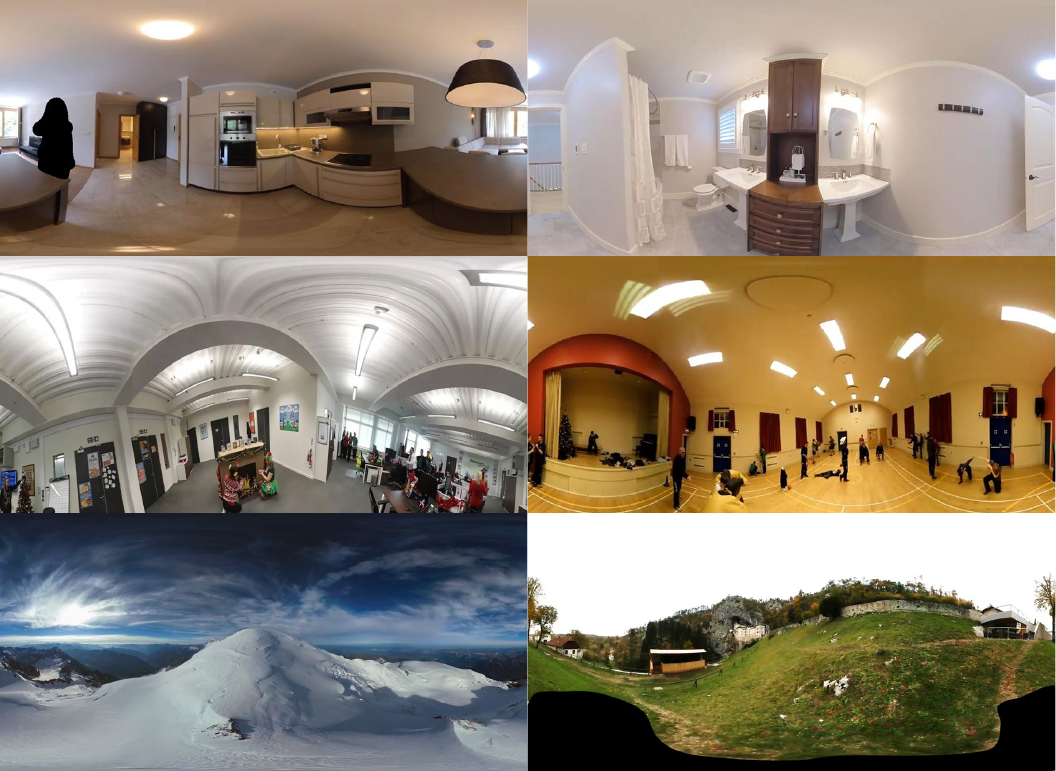}
    \caption{Sample omnidirectional images from the 360$^\circ$ in the Wild. Each row is grouped by category;~\textsc{Indoor}, \textsc{Outdoor}, and \textsc{Mannequin} challenge sequences, respectively. The person capturing the video is manually masked out, since the person (with a gimble or camera stick) consistently appears in the video, and it does not become relevant to the scene context. Images are cropped for the best view. Additional samples are available in the supplementary material.}
    \label{fig:data_category}
\end{figure}

After collecting the ``raw" videos, we detect drastic scene change~(\eg transition, fade in/out) using empirical gradient comparison between frames. From this strategy, we divide the videos into a few sub-sequences to feed appropriate input data to~\cite{opensfm}. Despite those keywords' restrictions, most of the videos suffer from the cameraman's presence in the field of view. Therefore, we masked it out utilizing Rotobrush2 in Adobe After Effects, as shown in~\Fig{data_category}. Additionally, we subdivide the videos into smaller sequences to simplify and speed up the reconstruction process.

\subsection{Statistics}

Our dataset contains 25K real images extracted from 273 videos manually downloaded from a video sharing platform. It exhibits various scenarios and environments that we classified into three distinct categories: \textsc{Indoor}, \textsc{Outdoor} and \textsc{Mannequin}. For each category, we provide a training, testing and validation set without overlap between sequences. The splits' statistics are available in \Tbl{split}. When compared with existing datasets for $360^\circ$ cameras (shown in~\Tbl{table_related_work}), our dataset is the only one providing a large quantity of real image along with their depth maps and cameras' poses. 

\begin{figure}[tb]
    \centering
    \includegraphics[width=.8\linewidth]{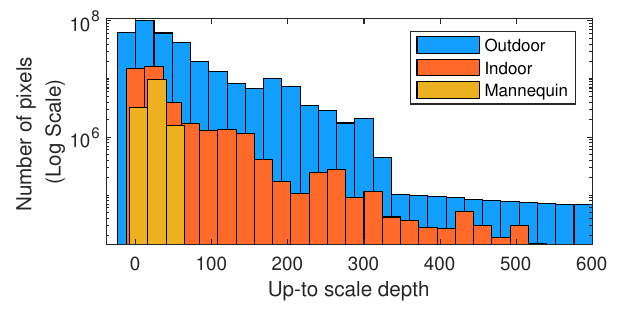}
    \caption{Statistics of depth values in our dataseet. Each category has different distribution because of the scene property. The number of pixels is displayed in log scale for better visualization.
    }
    \label{fig:distribution}
\end{figure}

To see the statistics of the reconstructed scenes, we examine the depth distribution. We calculated the depth histogram of each category by counting the number of pixels. We normalized the depth histogram using each video's mean depth because the ground-truth depth from internet videos has scale ambiguity. 
As shown in~\Fig{distribution}, the indoor scene has depth values relatively close to the camera, which is evident that indoor depth is restricted in the size of the building. Whereas in the outdoor image, the depth distribution in a far region is concentrated. 
Regarding the mannequin split, the narrow dynamic range can be justified by the very close shots of the human subjects in the scene.
Since our dataset has various depth distribution, we expect the dataset can contribute to the depth estimation approaches~\cite{Hu2019RevisitingSI,zioulis2019spherical}. Furthermore, view synthesis approaches such as NeRF~\cite{nerf} or NeRF++~\cite{nerfpp} could be trained on the diverse scenes concerning various depth distribution.


\begin{table}[]
\caption{Summary of depth dataset splits. The dataset is divided into the three categories, which represent the environment of the scene~(\eg \textsc{Indoor}, \textsc{Outdoor} and \textsc{Mannequin}). Note that \textsc{Mannequin} set contain both outdoor and indoor scene, including frozen people as like~\cite{frozen_data}.}
\label{tbl:split}
    \centering
    \resizebox{0.9\linewidth}{!}{
    \begin{tabular}{c|ccc|c}
    \hline
                                        &   Training &  Validation &  Test &  Total\\ \hline
    \multirow{1}{*}{ \textsc{Outdoor}}         & 2.1K  &  0.1K   & 0.2K  & 2.4K \\ \cline{2-5} \hline
    \multirow{1}{*}{ \textsc{Indoor}}        & 3.8K   & 0.2K   & 0.3K   & 4.3K\\ \cline{2-5} \hline
    \multirow{1}{*}{ \textsc{Mannequin}}      & 4.3K   & 0.1K   & 0.2K  & 4.6K\\ \hline
    
    \end{tabular}
    }
\vspace{2.2mm}

\end{table}

\subsection{Dataset Release}
Directly sharing the videos or image sequences is prohibited by the copyright law enforced on the video steaming platform -- from which the videos have been downloaded. To avoid any copyright infringement, we provide the link to each of these videos (as in~\cite{frozen_data}). Each video is decomposed in sub-sequences corresponding to different environments in the video (\ie different rooms in a house), the lists of frame indexes necessary to segment the video is available as a text file. For each frame, the following elements are provided: the depthmap, the binary mask (indicating moving objects in the scene) and the pose.


\section{Baselines}
In this section, we introduce the methods trained using our dataset for depth estimation and novel view synthesis.

\subsection{Depth Estimation}

For decades, estimating the 3D structure of the scene required multiple images acquired from different viewpoints. Recently, single image depth estimation~\cite{eigen2014depth,Hu2019RevisitingSI,fu2018deep,ramamonjisoa2019sharpnet} has been proposed for perspective cameras, but remains inadequate for images containing substantial geometric distortions like 360$^\circ$ images.

One of the seminal works focusing on single image depth for omnidirectional cameras is OmniDepth~\cite{omnidepth} in which synthetic $360^{\circ}$ images and their corresponding ground-truth depth maps is proposed. Omnidepth was extended to~\cite{zioulis2019spherical} which presents self-supervised learning using a stereo system. Alternatively, Payen~\etal~\cite{payen2018eliminating} approached depth estimation as a domain adaptation problem between perspective and unlabeled omnidirectional images with labels. More recently, BiFuse~\cite{bifuse} takes advantage of two different representations (namely, the cube map and the equirectangular images). To process the omnidirectional images directly, \cite{chen2020distortion} proposed deformable convolution, which is more appropriate to equirectangular images.
Despite compelling results, the previously mentioned methods are hardly applicable for ``in the wild" datasets due to the absence of metric and constant scale (each scene can exhibit its arbitrary scale reconstruction). This issue was recently addressed in MiDaS~\cite{Ranftl2020} which proposes to combine multiple datasets with different depth scales. To cope with this particular problem, MiDaS proposes a scale and shift-invariant loss. As a result of this dataset aggregation, MiDaS~\cite{Ranftl2020} demonstrates better performances in accuracy and generalization. 
Therefore, we decided to experiment using MiDaS~\cite{Ranftl2020} to demonstrate the applicability of our proposed dataset.

\subsection{Novel View Synthesis}


We are inspired by NeRF~\cite{nerf} to synthesize novel images from various viewpoints since it is currently state-of-the-art in this category. Since NeRF is intended for perspective cameras, we introduce our extension of NeRF for 360$^\circ$ images. We begin by introducing the key idea of NeRF and introduce our extension in this section.

NeRF represents a 3D scene as a neural radiance field, and the representation learned by a neural network that maps a 5D vector ($(x,y,z)$ location of a 3D point and its viewing direction~$(\theta,\phi)$) to RGB color values and opacity (4D vector). A set of rays that reached out from pixels in the image represents the radiance field, and the image can be made with the integration of the rays. NeRF samples 3D points on the ray to effectively estimate the field. For effective sampling, NeRF set lower and upper bounds in which the points are selected. However, the rendered image becomes problematic if the objects are located outside of the volume.

NeRF++~\cite{nerfpp} proposes a new inverted sphere parameterization for unbounded scenes. NeRF++ defines a sphere in which camera centers are located and normalizes the scene by making the sphere's radius to be 1. In this way, the scene is partitioned into two distinct volumes: the inner (foreground) and the outer (background) volumes. The two volumes are approximated via two separate NeRFs. The 3D points lie in the inner volume is sampled similarly to the vanilla NeRF. On the contrary, the 3D points outside the volume are sampled in an unbounded fashion $[0, \infty]$. To sample outer points $\mathbf{p}=[x,y,z]$, where $||\mathbf{p}||>1$, an homogeneous coordinate representation is proposed $[x',y',z',1/r]$, where $[x',y',z']$ is a unit vector along $[x,y,z]$ and $r$ is a value from which we sample in a range $[0,1]$. When $r$ goes to zero, the sampled points are near infinity.  A larger $r$ makes the point closer to the inner volume. 

\vspace{2mm} 
\noindent
\textbf{Proposed Nerf~\cite{nerfpp} extension for 360$^\circ$ images.}
NeRF~\cite{nerf} and NeRF++~\cite{nerfpp} achieve compelling performances, but these approaches have not been demonstrated for \emph{omnidirectional videos}. In this work, we introduce an extension for handling 360$^\circ$ images. 
Here, the two major components have to be considered. First, the rays have to be computed in compliance with the geometric model of $360^\circ$ images. Second, the bounds of the reconstruction volume have to be adjusted such that the entire scene can be taken into consideration.

Interestingly, the computation of the camera's rays for a full 360$^\circ$ image is trivial and does not require \emph{any calibration parameters}~\cite{bogdan2018deepcalib}.
Given a panorama $\mathcal{P}$ of size $H \times W$, each $i^{th}$ pixel $[p_x^i,p_y^i] \in \mathcal{P}$ can be expressed into spherical coordinates $[p_\theta^i,p_\rho^i]$ via a linear mapping:
\begin{equation}
p_{\theta}^i  =  \frac{2\pi p_{x}^i}{H} - \pi ~~~\mbox{and}~~~
p_{\rho}^i = -\frac{ \pi p_{y}^i}{W} + \frac{\pi}{2}.
\end{equation}
Finally, the rays' direction is expressed in Euclidean coordinates on the unit sphere which can be directly used in the NeRF and NeRF++ pipelines. The original NeRF implementation considers a bounded cost volume located in front of a reference camera. While this assumption is valid for perspective cameras, it is not compatible with omnidirectional vision systems.

A plausible solution would be to center the volume at the cameras' mean position. However, distant objects (located outside of this volume) would be badly reconstructed. Thus, we privileged the dual volumes strategy proposed by NeRF++, since it explicitly considers the entire surrounding scene by its spherical nature, and it also accounts for the faraway objects via an unbounded sampling technique.

We implement these ideas using NeRF++ and propose \emph{extended version of NeRF++}. In this paper, we use it as a baseline for view synthesis of omnidirectional images.


\begin{figure*}[htbp]
    \centering
    \begin{minipage}{.99\linewidth}
        
        \subfloat[Vanilla NeRF++]{
    	    \includegraphics[width=.32\linewidth]{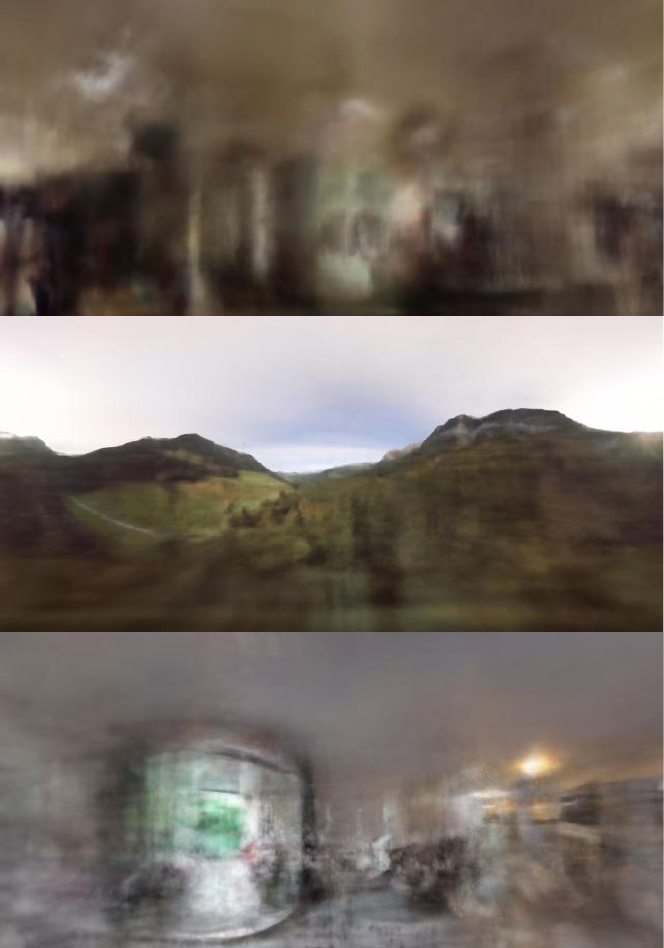}
    	} 
		\subfloat[Extended NeRF++]{
    	    \includegraphics[width=.32\linewidth]{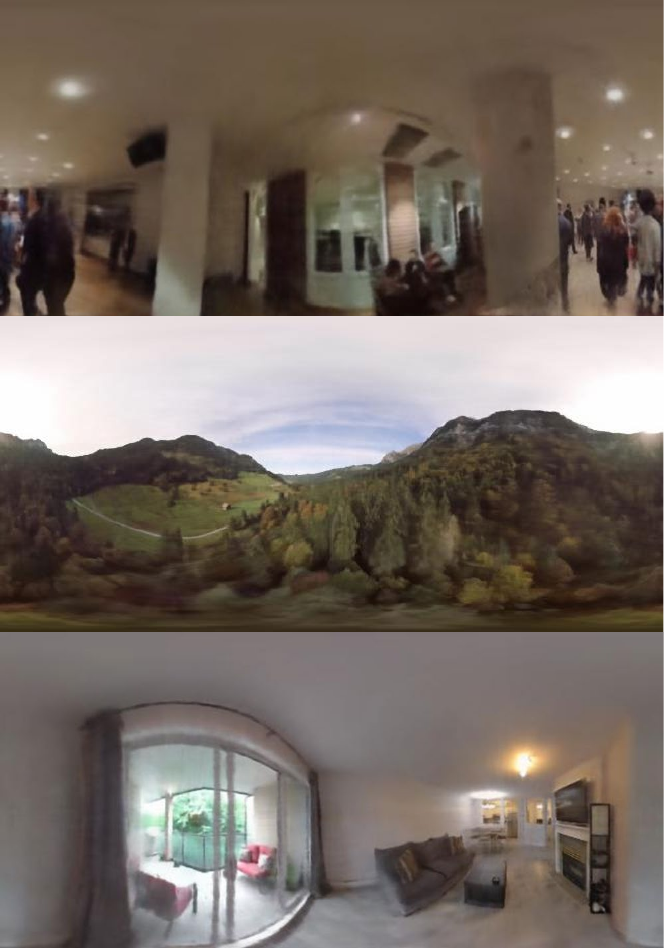}
    	}
	    \subfloat[Ground-truth]{
    	    \includegraphics[width=.32\linewidth]{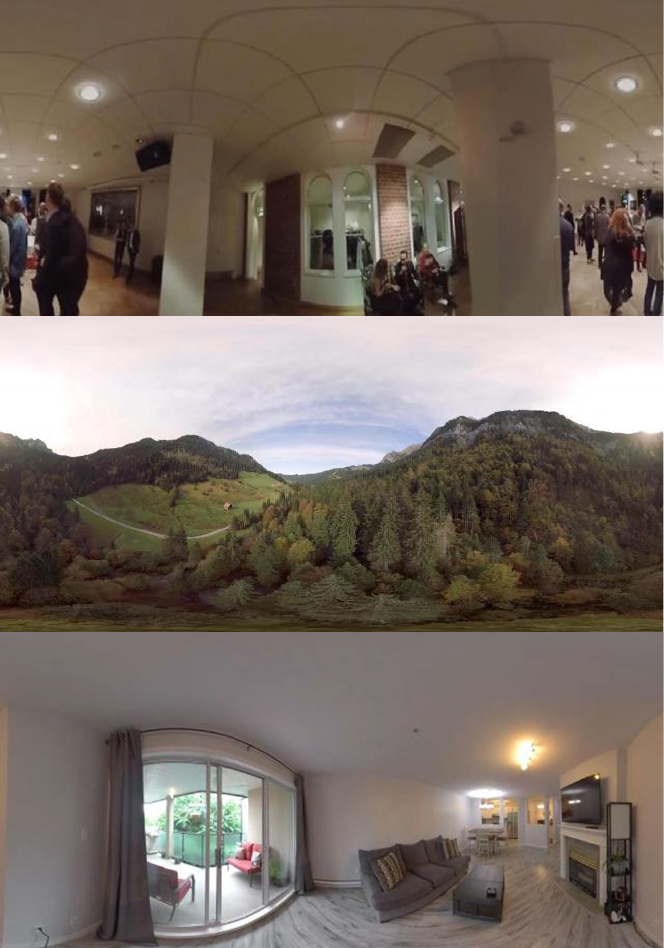}
    	}
	    \vspace{0.5mm}
    \end{minipage}
    \caption{Qualitative comparison results of 360$^\circ$ novel view synthesis. (a) Generated image with the official implementation of NeRF++~\cite{nerfpp}. Note that NeRF++ is not designed for 360$^\circ$ images and is included only to validate the necessity of our adaptation  (b) Results of extended NeRF++ for 360$^\circ$ images. (c) Ground-truth RGB images. Images are cropped for the best view. The images are taken from Mannequin-1 sequence~(The first row), Outdoor-1 sequence~(the second row) and Indoor-1 sequence~(the third row) in~\Tbl{nerfpp_and_ours}. Additional examples are available in the supplement.}
    \label{fig:qual_nerfpp_and_ours}
\end{figure*}

\section{Experiment}

To underline our dataset's relevance, we propose to evaluate two tasks: novel view synthesis and depth estimation.
Specifically, we trained NeRF++\cite{nerfpp} on different scene categories (i.e., \textsc{Indoor}, \textsc{Outdoor}, \textsc{Mannequin}) to evaluate image synthesis quality depending on the context.
On the other hand, we tested a depth estimation baseline with different training process including fine-tuning and training from the scratch.


\subsection{Implementation Details}

To train NeRF++ using our dataset, we follow the authors' recommendation regarding the hyperparameters' tuning. The optimization is conducted via an Adam optimizer with a learning rate of 0.0005, as described in~\cite{nerf}. We trained our models for a total of 8,000 iterations with eight Nvidia TITAN Xp GPUs. We resize the input image's resolution to have around 600 long edge pixels for better performances without changing the aspect ratio. We adopt the conventional metrics (PSNR, SSIM~\cite{wang2004image} and LPIPS~\cite{zhang2018unreasonable}) to assess the quality of our image synthesis.

Regarding the evaluation of single image depth, we fine-tune the networks presented in~\cite{Ranftl2020} on a single Quadro D8000 GPU. For training detail of~\cite{Ranftl2020}, the mini-batch size is set to 8 with a initial learning rate of 0.0005 during total 80 epoch. For the evaluation, we follow the basic metrics such as AbsRel~(mean absolute value of the relative error) and the proportional threshold $\delta = max(\frac{z}{z^{*}},\frac{z^{*}}{z}) < 1.25$ for measuring the depth quality which represented in~\cite{eigen2014depth}


\begin{figure*}[tb]    
    \centering
    \begin{minipage}{.995\linewidth}
		\subfloat[Original images]{
    	    \includegraphics[width=0.3245\linewidth]{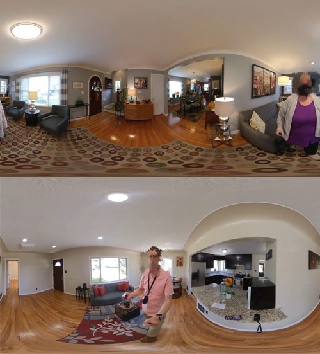}
    	    
    	}
    	\subfloat[Masked images]{
    	    \includegraphics[width=0.3245\linewidth]{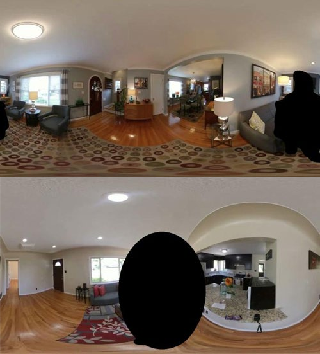}
    	}
    	\subfloat[Our predictions]{
    	    \includegraphics[width=0.3245\linewidth]{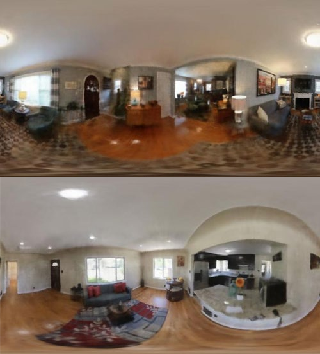}
    	}
	    \vspace{0.5mm}
    \end{minipage}
    \caption{360$^\circ$ view synthesis with masked videos. Extended NeRF++ that is trained on our dataset can generate images of the masked region. The network is trained with only a masked input image sequence. For better comparison, original images with a blurred human face are shown on the left. The images are taken from the Indoor-m1 sequence~(first row) and the Indoor-m2 sequence~(second row) in~\Tbl{nerfpp_and_ours}. Please see the supplement for the video.}
    \label{fig:mask_remove}
    \vspace{-2.5mm}
\end{figure*}

\begin{table}[]
\caption{Quantitative analysis of the 360$^\circ$ view synthesis from our NeRF++ for panormaic views. We test various environments, including images with masked regions of moving objects~(\eg, capturing person). We evaluate the results with PSNR, multi-scale structural similarity for SSIM~\cite{wang2004image}, LPIPS~\cite{zhang2018unreasonable}.}
\renewcommand{\arraystretch}{1.3}
\centering
\resizebox{0.7\linewidth}{!}{
\begin{tabular}{l||lll|}
 & $\uparrow$PSNR & $\uparrow$SSIM & $\downarrow$LPIPS \\ \cline{1-4} 
Mannequin-1 & 19.14 & 0.610 & 0.267 \\
Mannequin-2 & 16.72 & 0.539 & 0.429 \\
Outdoor-1 & 26.57 & 0.763 & 0.269 \\
Indoor-1 & 18.75 & 0.663 & 0.330 \\
Indoor-m1 & 15.88 & 0.497 & 0.442 \\
Indoor-m2 & 14.46 & 0.564 & 0.421
\end{tabular}%
}
\vspace{1mm}

\label{tbl:nerfpp_and_ours}
\end{table}

\subsection{Results}

\begin{figure*}[tb]
        \centering
		\subfloat[input images]{
    	    \includegraphics[width=0.246\linewidth]{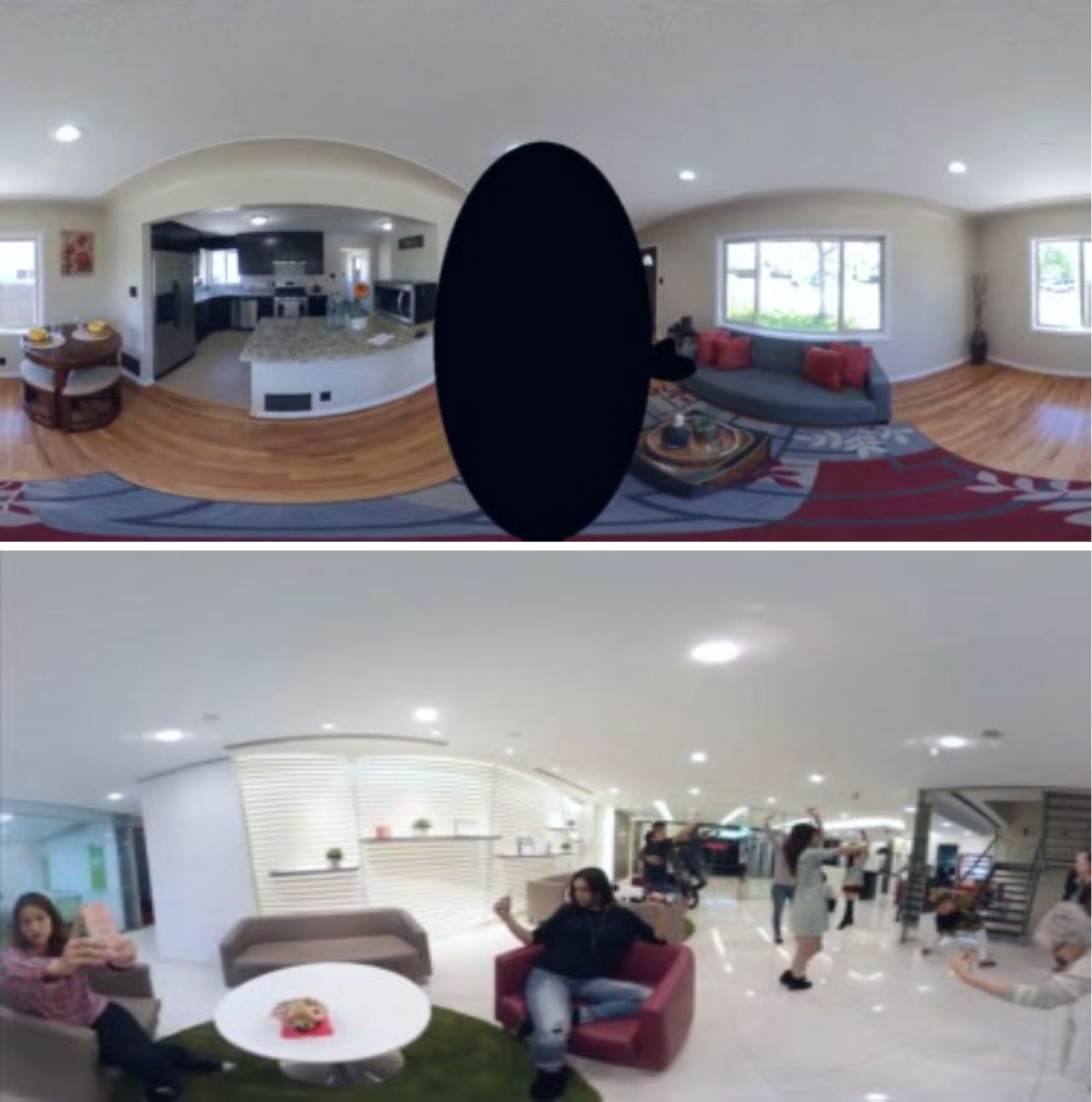}
    	}
    	\subfloat[Ground-truth]{
    	    \includegraphics[width=0.246\linewidth]{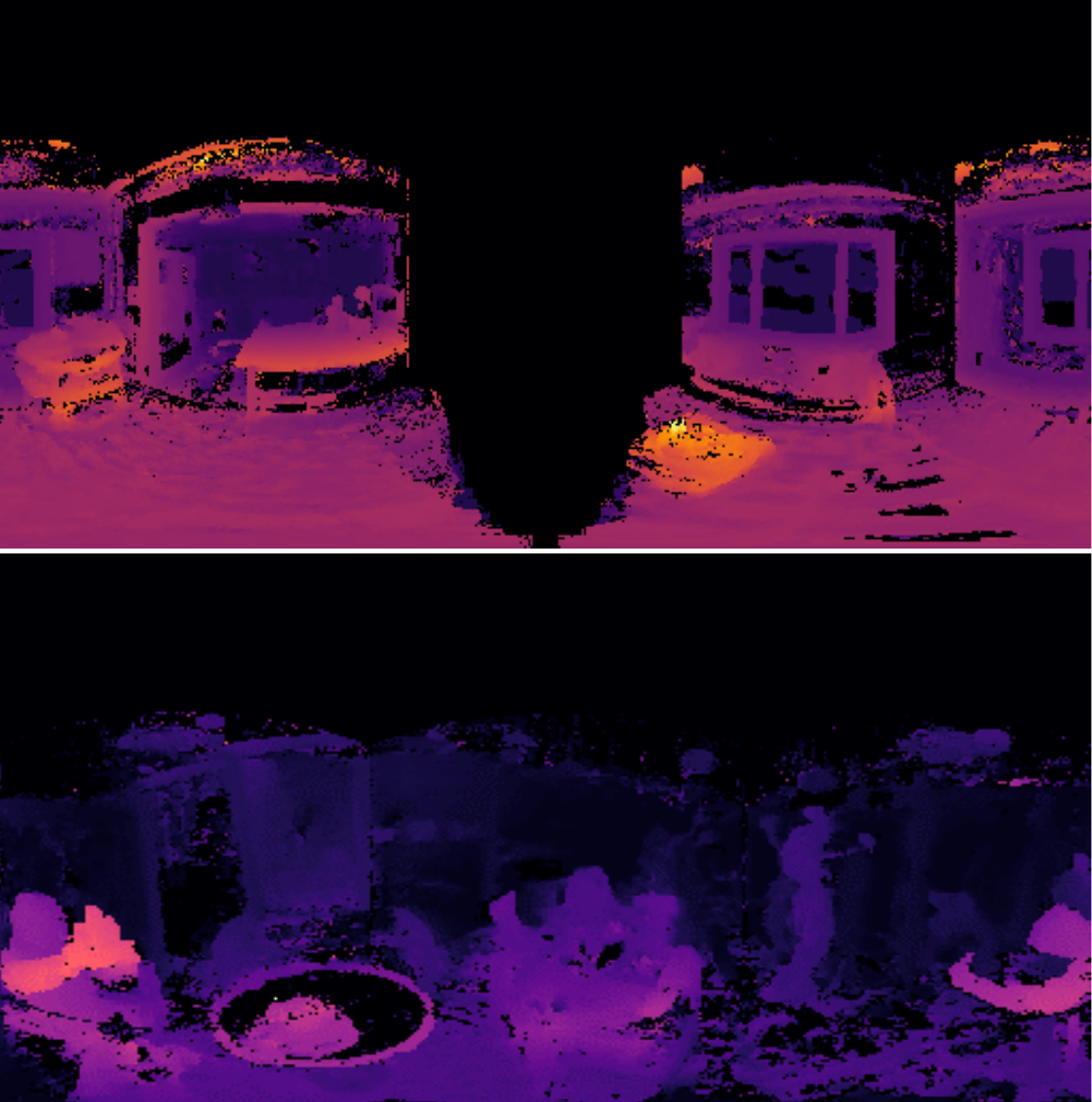}
    	}
    	\subfloat[MiDaS-Sc~\cite{Ranftl2020}]{
    	    \includegraphics[width=0.246\linewidth]{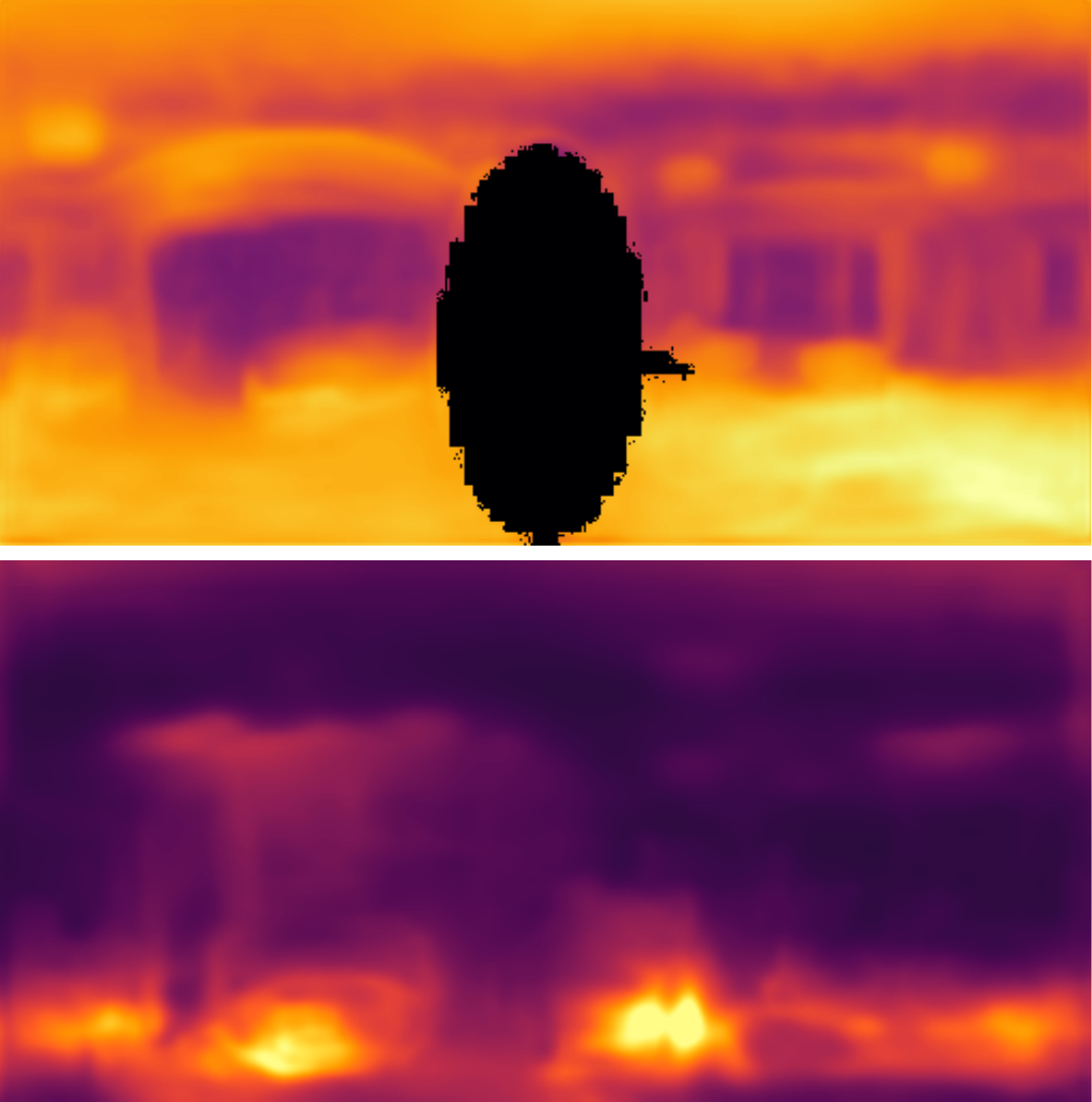}
    	}
	    \vspace{1.5mm}

    \vspace{2mm}
    \caption{Qualitative comparison of depth estimation approaches. Note that these results have been computed on unseen test internet videos. Notice that the first row depicts a video with a moving person that has been manually masked. Despite this distraction the network sucessfully estimates the depth of the scene.} 
    \label{fig:qual_depth}
\end{figure*}

\vspace{1mm} \noindent \textbf{Depth Estimation.} 
We evaluate single-view depth estimation approaches from a subset of 11k image-depth pairs. Among these 360$^\circ$ images, we randomly selected 700 images (from different scenes than the training set) as a test set. In this experiment, we compare Ranftl~\etal~\cite{Ranftl2020} with different training variant~(from the scratch and fine-tuning) on our data. Ranftl~\etal propose a robust single-view depth estimation across different datasets~(various scale). It is, therefore, a solution of choice to test our dataset. 
The obtained results are available in Table~\Tbl{depth_score}.
Note that these results have been obtained from our own implementation of MiDaS~\cite{Ranftl2020} since their code is only partly publicly available.
For the sake of completness, we proposed to run three configurations of MiDaS~\cite{Ranftl2020}, a model pretrained on the 3D60 dataset~\cite{zioulis2019spherical}, a fine-tuned version of this model on our data, and, finally MiDaS trained from scratched on our dataset. The fine-tuned network demonstrates the best accuracy closely followed by the training from scratch on our data. We conjecture that the pretraining allows a faster convergence and lead to higher performances, however, our dataset in itself is sufficient to ensure a reliable depth estimation. 
Since MiDaS has not initially been developed for omnidirectional cameras, we can notice that the depth quality is relatively low when compared to perspective image results. Thus, pursuing research toward novel approaches for scale-invariant omnidirectional depth estimation seems to be a relevant direction in which our dataset can be useful.
Despite the incomplete nature of the depth map used for training (as in~\cite{frozen_data}), both network predicts dense depth as demonstrated in outdoor and indoor environments, as shown in the~\Fig{qual_depth}.


 \noindent \textbf{Novel view synthesis.} 
To demonstrate the relevance of our dataset in the context of novel view synthesis, we propose to apply our adapted NeRF++ implementation to a few sequences. Specifically, we picked six different videos to perform this task: two mannequins, one outdoor, and three indoor. In non-mannequin videos, the presence of moving objects (often pedestrians) contaminate the data. Thus these moving objects have been manually masked as shown in~\Fig{mask_remove}. In our test set, two sequences (namely Indoor-m1 and Indoor-m2 in~\Tbl{nerfpp_and_ours}) contain such masking. 


\begin{figure}[tb]
    \centering
    \includegraphics[width=0.7\linewidth]{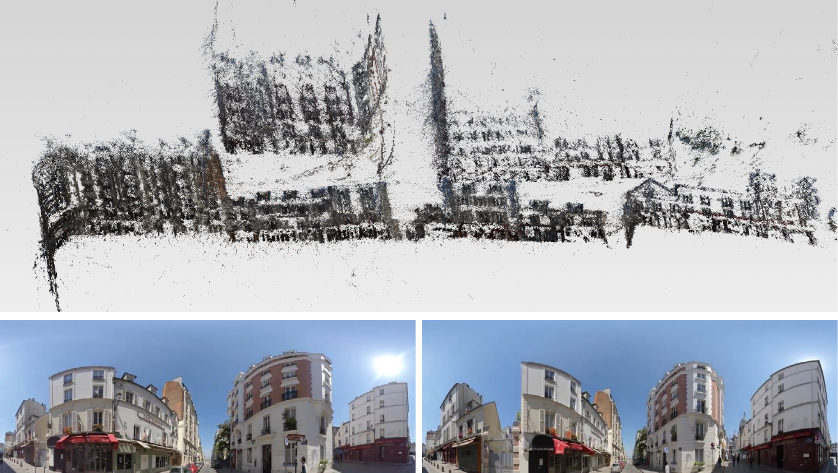}
\caption{3D Reconstruction by using proposed dataset. Reliable ground-truth is achieved by OpenSfM and COLMAP pipelines. }
    \label{fig:recon}
\end{figure}
\vspace{-1.5mm}

A set of representative qualitative results is available in~\Fig{qual_nerfpp_and_ours}. 
To underline the necessity of our NeRF++ extension to 360$^\circ$ images, we additionally included results from the original \emph{Vanilla} NeRF++ with arbitrary camera's parameters. Note that this original approach cannot truly perform on this kind of image since the equation of the ray is not appropriated (regardless of the perspective camera's parameters).
Despite the complex nature of $360^{\circ}$ images, our adaptation allows us to reconstruct novel views effectively. To provide a more in-depth analysis, we use the quantitative image quality metrics in~\Tbl{nerfpp_and_ours}.


For real images with high resolution, the original NeRF++ paper~\cite{nerf} reports the following average metrics: 20.19, 0.82, 0.32 for PSRN, SSIM and LPIPS respectively in perspective camera. Due to the complexity of this task for $360^{\circ}$ images, our synthesised image quality is lower than~\cite{nerf} which suggests that large improvements for panoramic view synthetic can still be achieved. To develop novel algorithms dedicated to omnidirectional cameras, this dataset of posed images certainly represents an important resource.



Additionally, we tested our NeRF++ on \emph{masked videos}, as shown in~\Fig{mask_remove}. For the masked pixels, the training ray of the masked area is disregarded. Thus, the model naturally learns a static scene from multi-view input images without any disrupted moving object. Surprisingly, it synthesizes the image of the masked scene, as shown in (b)~\Fig{mask_remove}. It demonstrates that the model implicitly learns the structure of the environment using partial observations. The implicit 3D scene knowledge from the unmasked area is propagated to other relevant scene structure, enabling the masked region gradually learn the shared scene structures.

\begin{table}
\caption{Comparison of single-view depth estimation approaches. Sc~(from the scratch) and Ft~(fine-tuning) denote evaluation with a training from the scratch and a fine-tuning process based on  MiDaS~\cite{Ranftl2020}, respectively. MiDaS-3D60 indicates network pretrained on 3D60 dataset. This omnidirectional indoor dataset has been captured from depth sensors~\cite{zioulis2019spherical}.}
\renewcommand{\arraystretch}{1.3}
\centering
\resizebox{0.95\linewidth}{!}{
\setlength \tabcolsep{1.2pt}

\begin{tabular}{cc cc cc cc cc cc}
\hline
\multirow{2}{*}{ } & &$\uparrow$ $\delta < 1.25$ &   &    &$\uparrow$ $\delta < 1.25^2$ &  &  $\uparrow$ $\delta < 1.25^3$ &  & $\downarrow$ AbsRel\\
\hline

MiDaS-3D60  &     & 0.650 &    &    & 0.755   &    &     0.833   & &     0.684   &  \\

MiDaS-3D60-Ft  &     & \bfseries0.716    &    &    & \bfseries0.823 &    &    \bfseries 0.885   & &    \bfseries 0.544   &  \\

MiDaS-Sc~\cite{Ranftl2020}  &     & 0.701  &    &    & 0.819 &    &    \bfseries 0.885 & &    0.557 &  \\



\hline
\end{tabular}
}
\vspace{1.3mm}

\label{tbl:depth_score}
\end{table}


\section{Conclusion}

We demonstrated an omnidirectional video dataset for depth estimation and novel view synthesis, exploiting diverse internet videos on YouTube. We shows that our dataset enable each task to learn the various image categories~(including \textsc{Indoor}, \textsc{Outdoor} and \textsc{Mannequin}). 
We will provide the dataset to the public, and the dataset contains video links of each scene, data split information, ground-truth from 3D reconstruction, and so on. 

Our dataset has a limitation that the SfM pipeline cannot perfectly reconstruct the scene, and calculated depth is not the metric scale, as normally like photogrammetry pipelines with unknown camera intrinsics. 
Our future work includes extending the tasks of the dataset, such as neural point cloud rendering~\cite{Dai_2019_neuralpointcloudrendering}. Overall, we believe that 360$^\circ$ in the Wild contribute to the community with new omnidirectional video data that resolve the real-world scenarios.

\balance
{\small
\bibliographystyle{ieee_fullname}
\bibliography{bare_conf}
}

\end{document}